\documentclass[10pt,letterpaper,twoside,twocolumn,journal,final]{IEEEtran}
\usepackage[utf8]{inputenc}
\usepackage[T1]{fontenc}
\usepackage{amssymb}
\usepackage{amsmath}
\usepackage{amsfonts}
\usepackage{nicefrac}
\usepackage{subfig}
\usepackage{graphicx}
\usepackage{microtype}
\usepackage{booktabs}
\usepackage[table]{xcolor}
\usepackage{siunitx}
\usepackage{paralist}
\usepackage{enumitem}
\usepackage[ruled,vlined,linesnumbered]{algorithm2e}
\usepackage[noadjust]{cite}
\usepackage{url}
\usepackage{tikz}
\usetikzlibrary{matrix,shapes,decorations.markings,arrows,calc}
\graphicspath{{figs/}}
\makeatletter
\newcommand{\removelatexerror}{\let\@latex@error\@gobble}
\def\input@path{{figs/}}
\makeatother
\DeclareMathOperator*{\argmax}{arg\,max}
 
\begin{document}

\title{Deep Learning with a Classifier System: \\Initial Results}

\author{Richard~J.~Preen, and Larry~Bull%
	\thanks{Manuscript date of current version March 1, 2021.}%
	\thanks{R.~J.~Preen and L.~Bull are with the Department of Computer Science and Creative Technologies, University of the West of England, Bristol BS16 1QY, UK (e-mail: richard2.preen@uwe.ac.uk; larry.bull@uwe.ac.uk).}%
    \thanks{}%
}
 
\bstctlcite{IEEEexample:BSTcontrol}
\IEEEpubid{}%
\markboth{}%
{PREEN \MakeLowercase{and} BULL:~DEEP LEARNING WITH A CLASSIFIER SYSTEM}
 
\maketitle

\begin{abstract}
This article presents the first results from using a learning classifier system capable of performing adaptive computation with deep neural networks. Individual classifiers within the population are composed of two neural networks. The first acts as a gating or guarding component, which enables the conditional computation of an associated deep neural network on a per instance basis. Self-adaptive mutation is applied upon reproduction and prediction networks are refined with stochastic gradient descent during lifetime learning. The use of fully-connected and convolutional layers are evaluated on handwritten digit recognition tasks where evolution adapts
\begin{inparaenum}[(i)]
    \item the gradient descent learning rate applied to each layer
    \item the number of units within each layer, i.e., the number of fully-connected neurons and the number of convolutional kernel filters
    \item the connectivity of each layer, i.e., whether each weight is active
    \item the weight magnitudes, enabling escape from local optima.
\end{inparaenum}%
The system automatically reduces the number of weights and units while maintaining performance after achieving a maximum prediction error.
\end{abstract}


\section{Introduction}

\IEEEPARstart{S}{ystems} that learn to exploit deep feature hierarchies have significantly contributed to improvements in the current state-of-the-art in many application domains, including speech recognition, object detection, natural language processing, and reinforcement learning~\cite{LeCun:2015}. As these systems continue to be applied to increasingly complex problems, the ability to hand-craft efficient architectures is becoming more difficult. This is particularly true of dynamic systems, which perform adaptive computation in response to the input~\cite{Han:2021}.

A growing number of algorithms have therefore been proposed to automatically search and optimise large and complex neural systems~\cite{Elsken:2019}. Evolutionary algorithms (EAs) have proven to be flexible and successful in this endeavour, both in the more restricted case of exploring novel architectures~\cite{Liu:2020}, and in the wider optimisation of parameters, activation functions, and objectives, etc.~\cite{Stanley:2019}.

As the computational expense and wall-clock time involved in the training and evaluation of these systems increases with the size of the model and data set, techniques for reducing these costs have become increasingly important. This is a particular concern for population-based approaches such as EAs, which can involve hundreds or thousands of large models. Encouraging model sparsity and pruning redundancy can result in significant reductions in this cost, reducing the overall size while maintaining performance~\cite{Hoefler:2021}. The use of surrogate-modelling is a long established technique to reduce the number of costly EA fitness evaluations, and has also been used to estimate the validation error of candidate neural networks~\cite{Zhang:2021}.

Current state-of-the-art approaches have typically designed large global networks whose architecture and parameters remain unchanging to varying input~\cite{Liu:2020}. However, a number of benefits in terms of accuracy and efficiency can be achieved by allowing networks to dynamically adapt their structure and parameters~\cite{Han:2021}. For example, enabling dynamic depth, width, routing, and pruning. These can be performed on a per instance basis, or in response to the spatial positioning or temporal dimension of the input. 

The learning classifier system (LCS) XCSF~\cite{Wilson:2001} is a framework for performing adaptive computation in response to the input. XCSF achieves this through a population-based approach to conditional computation~\cite{Bengio:2016} wherein individual gating/guarding components are each associated with a local approximation. It therefore maintains the ability to decompose large problems into a set of smaller local solutions that collectively cover the global domain. Each sub-solution is activated only for the subset of instances that its condition permits. Error is therefore allocated directly to the components responsible, potentially enabling faster convergence and reduced computational cost~\cite{Preen:2019}.

Shallow neural networks have been widely explored within LCS for supervised and reinforcement learning tasks~\cite{Bull:2002,Bull:2003,OHara:2005a,OHara:2005b,Lanzi:2006,Lanzi:2007b,OHara:2007,Dam:2008,Sancho-Asensio:2014}. More recently, they have been used in the unsupervised learning of an ensemble of heterogeneous autoencoders~\cite{Preen:2019}. Spiking neural networks~\cite{Howard:2016} and cyclic graphs~\cite{Preen:2013,Preen:2014}, which are capable of deep architectures have also been explored, although these do not include the ability for refinement with local search.

The combination of state-of-the-art deep learning techniques with LCS is now emerging as a promising area of research. Deep neural networks have been used to extract features and reduce the dimensionality of data presented to XCS for classification~\cite{Matsumoto:2018,Tadokoro:2019}. Additionally, deep convolutional networks have been used within Pittsburgh-style LCS for detecting database intrusion~\cite{Kim:2019,Bu:2020}. While Pittsburgh LCS are rule-based systems, each chromosome represents a complete global solution to the problem, similar to standard EAs.

\IEEEpubidadjcol

To date, use of deep hierarchical structures in conjunction with stochastic gradient descent directly within a Michigan-style LCS capable of performing adaptive computation in response to the input remains unexplored. This article extends XCSF to perform deep learning for the first time. We explore the performance of fully-connected and convolutional neural networks where the number of units as well as the connectivity evolves, i.e., deep heterogeneous niched networks may emerge. Furthermore, as in \cite{Preen:2019}, we employ a self-adaptive scheme wherein the rate of mutation is continually adapted throughout the search process. Each layer adapts to a localised rate of gradient descent.

In our initial results, we present the performance of XCSF with deep neural networks on two popular handwritten image recognition tasks. \cite{Frey:1991} was the first to use LCS in this endeavour, however producing mediocre results. More recently, \cite{Ebadi:2012,Ebadi:2014} examined the use of XCS and UCS with Haar-like features for image recognition on the \textsc{mnist digits} data set. While the approach created a more human-interpretable rule set than neural networks, the classification accuracy reported was not competitive with the current state-of-the-art.



\section{Methodology}
\label{section:methodology}

\subsection{Overview}

XCSF~\cite{Wilson:2001} is an accuracy-based online evolutionary machine learning system with locally approximating functions that compute classifier payoff prediction directly from the input state. XCSF attempts to find solutions that are accurate and maximally general over the global input space. However, it can automatically subdivide the input space into simpler local approximations through an adaptive niching mechanism.

XCSF is rule-based and maintains a population of classifiers, where each classifier $cl$ consists of
\begin{inparaenum}[(i)]
    \item a condition component $cl.C$ that determines whether the rule matches input $\vec{x}$
    \item an action component $cl.A$ that selects an action $a$ to be performed for a given $\vec{x}$
    \item a prediction component $cl.P$ that computes the expected payoff for performing $a$ upon receipt of $\vec{x}$.
\end{inparaenum}%
XCSF thus generates rules of the general form: \texttt{IF} \textit{matches} $\leftarrow cl.C(\vec{x})$ \texttt{THEN} perform action $a \leftarrow cl.A(\vec{x})$ and \texttt{EXPECT} payoff $\vec{p} \leftarrow cl.P(\vec{x})$.

Following \cite{Preen:2019}, here XCSF maintains a population of classifiers where each $cl.C$ and $cl.P$ is a separate neural network. A population set $[P]$ of $N=500$ classifiers are initialised randomly and undertake Lamarckian learning. That is, after the application of evolutionary operators to both $cl.C$ and $cl.P$ during reproduction, stochastic gradient descent updates $cl.P$ during reinforcement. The resulting $cl.P$ weights are copied to offspring upon parental selection.

The initial weights of each network in $[P]$ are set to small random values sampled from a Gaussian normal distribution with mean $m=0$ and standard deviation $\sigma=0.1$. Biases are zero initialised. Each $cl.C$ output layer contains a single neuron with a linear activation function that determines whether the rule matches a given input $\vec{x}$. That is, upon receipt of $\vec{x}$, a match set $[M]$ is formed by adding all $cl \in [P]$ whose $cl.C$ outputs a value greater than 0.5. If there are no matching classifiers for a given input, a covering mechanism is invoked wherein networks with random weights and biases are generated by the same initialisation method until the current input is matched, however using a larger $\sigma=1$. 

Data labels representing the category/class associated with the sample instances are one-hot encoded as the target vector $\vec{y}$. Each $cl.P$ output layer $\vec{p}$ uses a softmax activation function and contains as many output neurons as there are label categories. A loss function $\mathcal{L}$ computes the classification error for each individual classifier by comparing the most likely class $\argmax \vec{p}$ with $\argmax \vec{y}$. For system output, the $[M]$ fitness-weighted average prediction is computed as is usual in XCSF, i.e., $\vec{p} = \nicefrac{\sum_{cl \in [M]} cl.F \cdot cl.\vec{p}}{\sum_{cl \in [M]} cl.F}$, with the most likely class selected, as with individual classifiers.

Classifier updates and the EA take place as usual within $[M]$. That is, each classifier $cl_j \in [M]$ has its experience $exp$ incremented and fitness $F$, error $\epsilon$, and set size $as$ updated using the Widrow-Hoff delta rule with learning rate $\beta \in [0,1]$ as follows.
\begin{itemize}[label=$\triangleright$]
    \item Error: $\epsilon_j \leftarrow \epsilon_j + \beta(\mathcal{L}(\vec{p}_j, \vec{y}) - \epsilon_j)$ \\
    \item Accuracy: $\kappa_j = 
        \begin{cases}
            1 & \text{if $\epsilon_j < \epsilon_0$} \\
            {\alpha(\nicefrac{\epsilon_j}{\epsilon_0})}^{-\nu} & \text{otherwise}.
        \end{cases} $ \\
        With target error threshold $\epsilon_0$ and accuracy fall-off rate $\alpha \in [0,1]$, $\nu \in \mathbb{N}_{>0}$.
    \item Relative accuracy: $\kappa_j' = \nicefrac{\kappa_j \cdot num_j}{\sum_j \kappa_j \cdot num_j}$ \\
    Where classifier numerosity initialised $num=1$.
    \item Fitness: $F_j \leftarrow F_j + \beta(\kappa_j'-F_j)$
    \item Set size estimate: $as_j \leftarrow as_j + \beta(|[A]|-as_j)$
\end{itemize}

Subsequently, each $cl.P$ within $[M]$ is updated using simple stochastic gradient descent~\cite{Rumelhart:1986} with a layer-specific learning rate $\eta \in \mathbb{R}_{>0}$ and momentum $\omega \in [0,1]$. That is, the chain rule is applied at match time $t$ to compute the partial derivative of the error with respect to each weight $\nicefrac{\partial \mathcal{E}}{\partial w}$, and the weight change:
\begin{equation}
\Delta w_t = -\eta \nicefrac{\partial \mathcal{E}}{\partial w_t} + \omega \Delta w_{t-1}
\end{equation}
Gradient descent is not applied to $cl.C$. 

The EA is applied to classifiers within $[M]$ if the average set time since its previous execution exceeds $\theta_{\text{EA}}$. Upon invocation, the time stamp $ts$ of each classifier is updated. Two parents are chosen based on their fitness via roulette wheel selection and $\lambda$ number of offspring are created using the evolutionary operators as described below. Offspring fitness and error are inherited from the parents and discounted by reduction parameters for error $\epsilon_R$ and fitness $F_R$. Offspring experience $exp$ and numerosity $num$ are set to 1. The resulting offspring are added to $[P]$ and the maximum population size $N$ is enforced. Since the possibility exists that a $cl.C$ (not generated through covering) may never match any inputs, any classifiers that have not matched any inputs within 10000 trials since creation are prioritised for removal. If none are found, classifiers are selected via roulette wheel with the deletion vote as usual.

\begin{figure}[t]
	\centering
	\small
    \begin{tikzpicture}
        \node(p) at (-2.5,0)[align=center, rectangle, rounded corners=5pt, draw, thick, minimum width=25mm, minimum height=10mm] {$[P]$\\Population};
        \node(m) at (2.5,0)[align=center, rectangle, rounded corners=5pt, draw, thick, minimum width=25mm, minimum height=10mm] {$[M]$\\Match Set};
        \node(env) at (0,2.1)[align=center, rectangle, rounded corners=0pt, draw, thick, minimum width=70mm, minimum height=8mm] {{\em Environment}};
        \node(ea) at (0,-3.3)[circle, draw, dashed, thick, minimum height=12mm] {EA};
        \node(c) at (0,-1.75)[circle, draw, dashed, thick, minimum height=12mm] {Cover};
        \node(target) at (3.6,1.1)[] {target $\vec{y}$};
        \node(match) at (0,0.2)[] {match};
        \node(input) at (-3.1,1.1)[] {input $\vec{x}$};
        \node(reproduce) at (1.7,-3)[] {reproduce}; 
        \node(delete) at (-1.4,-3)[rotate=0] {delete};
        \node(insert) at (-1.1,-1.1)[] {insert}; 
        \node(syspred) at (1.05,1.1) [align=left] {prediction $\vec{p}$};
        \node(update) at (3.,-2.)[] {update};
        \draw [black, line width=1pt, ->, >=stealth] (p) edge node [above] {} (m);
        \draw [black, line width=1pt, ->, >=stealth] (p.north|-env.south) -- (p.north);
        \draw [black, line width=1pt, ->, >=stealth] ([xshift=-5mm]m.north) -- ([xshift=-5mm]m.north|-env.south);
        \draw [black, line width=1pt, ->, >=stealth] ([xshift=5mm]m.north|-env.south) -- ([xshift=5mm]m.north) ;
        \draw [black, dashed, line width=1pt, ->, >=stealth]
        (m) edge [bend left=20] node [above] {} (c)
        (c) edge [bend left=20] node [above] {} (p)
        (ea) edge [bend right=30] node [above] {} (m)
        (ea) edge [bend left=30] node [above] {} (p);
        \draw [black, dashed, line width=1pt, ->, >=stealth] (update) edge node [above] {} (m.south-|update.north);
    \end{tikzpicture}	
    \caption{Schematic illustration of XCSF for supervised learning.}
	\label{fig:xcsf}
\end{figure}
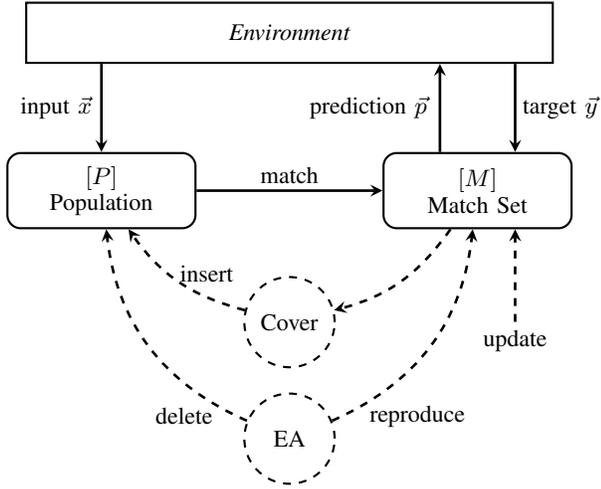

The deletion vote for each $cl \in [P]$ is set proportionally to the match set size estimate $as$. However, the vote is increased by a factor $\nicefrac{\overline{F}}{F_j}$ for classifiers that are sufficiently experienced ($exp_j > \theta_{\text{del}}$) and with small fitness $F_j < \delta \overline{F}$; where $\overline{F}$ is the $[P]$ mean fitness, and $\delta=0.1$. Classifiers selected for deletion have their $num$ decremented, and in the event that $num<1$ are removed from $[P]$. A schematic overview is shown in Fig.~\ref{fig:xcsf}. See \cite{Butz:2006} for a detailed introduction to XCSF.

\subsection{Evolutionary Operators}

Following~\cite{Preen:2019}, crossover is omitted from the EA and a layer-specific self-adaptive mutation scheme is used. That is, the mutation rates are locally evolving entities, continuously adapting throughout the search process. In this article however, we adopt the rate selection adaptation method from \cite{Serpell:2010}. Each layer within each classifier therefore maintains a vector of mutation rates $\ell.\vec{\mu}$ initially selected randomly from a set $\mathcal{M}$ of 10 possible values. These parameters are passed from parent to offspring, however rates are randomly re-selected from the set with 10\% probability. This method has been shown to be more robust for combinatorial optimisation and allows the rates to more easily increase from small values to escape local optima~\cite{Serpell:2010}. Self-adaptive mutation has previously been shown to reduce the number of hand-tunable parameters and improve performance. Similar to \cite{Preen:2019}, here four types of mutation are explored such that for each layer:

\begin{itemize}
    \item The rate of gradient descent is a locally evolving entity. Each layer maintains its own $\eta$. These values are constrained [$10^{-4},0.01$] and seeded uniformly random. A self-adaptive mutation rate controls the $\sigma$ of a random Gaussian added to each $\eta$.
    \item A second self-adaptive rate controls the number of hidden units to add or remove, i.e., the number of neurons for fully-connected layers and kernel filters for convolutional layers. When triggered, a random number of units are added or removed, drawn from a Gaussian distribution $\sim \mathcal{N}_{\mathbb{N}_{\neq0}}[-h_M,h_M]$ where the parameter $h_M$ determines the maximum number of units that may be added or removed per mutation event.
    \item The connectivity is mutated through the use of two additional self-adaptive rates. The first rate controls the probability of enabling currently disabled connections, and the second controls the rate of disabling currently enabled connections. Networks are initialised fully-connected. When a connection is disabled, the corresponding weight value is set to zero and is excluded from mutation and gradient descent updates. Upon activation, the weight is set to a small random value $\sim \mathcal{N}(0,0.1)$.
    \item The magnitude of weights and biases are adapted through the use of a single self-adaptive mutation rate, which controls the $\sigma$ of a random Gaussian added to each weight and bias.
\end{itemize}

\begin{figure}[t]
    \removelatexerror
    \begin{algorithm}[H]
    	\SetNoFillComment
    	\small
    	\DontPrintSemicolon
        \SetProcNameSty{textsc}
        \SetProcArgSty{textsc}
	    \SetKwProg{myalg}{Algorithm}{}{}
	    \SetKwProg{myproc}{procedure}{}{}
	    \SetKwFunction{roulette}{SelectParentsRouletteFit}
	    \SetKwFunction{sam}{AdaptMutationRates}
    	\SetKwFunction{sgd}{MutateSGDRate}
    	\SetKwFunction{units}{MutateUnits}
    	\SetKwFunction{connect}{MutateConnectivity}
    	\SetKwFunction{weights}{MutateWeights}
    	$\mathcal{M} = \{.0005, .001, .002, .003, .005, .01, .015, .02, .05, .1\}$\;
    	$\{c1p, c2p\} \in [M] \leftarrow $ \roulette()\;
    	\For{$\nicefrac{\lambda}{2}$ number of times} {
    		$\{c1, c2\} \leftarrow Copy(c1p, c2p)$\;
    	    \For{$cl \in \{c1, c2\}$} {
    	        \For{layer $\ell \in \{cl.C, cl.P\}$} {
    	            \sam($\ell$)\;
    	            \sgd($\ell$)\;
    	            \units($\ell$)\;
    	            \connect($\ell$)\;
    	            \weights($\ell$)\;
    	        }
            }
        }
	    \myproc{\sam{$\ell$}}{
            \For{$\mu_i \in \ell.\vec{\mu}$} {
                \If{$U[0,1] < 0.1$} {
                    $\mu_i \xleftarrow{U} \mathcal{M}$\;
                } 
            }
	    }
	    \myproc{\sgd{$\ell$}}{
            $\ell.\eta \leftarrow \ell.\eta + \mathcal{N}(0,\ell.\mu_1)$\;
	    }
	    \myproc{\units{$\ell$}}{
            \If{$U[0,0.1] < \ell.\mu_2$} {
                $\ell.h \leftarrow \ell.h + \mathcal{N}_{\mathbb{N}_{\neq0}}[-h_M,h_M]$\;
            } 
	    }
	    \myproc{\connect{$\ell$}}{
            \For{$w_i \in \ell.\vec{w}$} {
                \If{$a_i$ is disabled \textbf{and} $U[0,1] < \ell.\mu_3$} {
                    $w_i, a_i \leftarrow \mathcal{N}(0,0.1)$, enabled\;
                } 
                \ElseIf{$a_i$ is enabled \textbf{and} $U[0,1] < \ell.\mu_4$} {
                    $w_i, a_i \leftarrow 0$, disabled\;
                }
            }
	    }
	    \myproc{\weights{$\ell$}}{
            \For{$w_i \in \ell.\vec{w}$} {
                \If{$a_i$ is enabled} {
                    $w_i \leftarrow w_i + \mathcal{N}(0,\ell.\mu_5)$\;
                }
            }
	    }
    	\caption{Offspring Creation.}
    	\label{alg}
    \end{algorithm}
\end{figure}

Algorithm~\ref{alg} presents an outline of the process for offspring creation. Pressure to evolve minimally sized networks is achieved by further altering the population size enforcement mechanism as follows. A moving average of the system error is tracked using the same $\beta$ update for individual classifiers. Each time a classifier must be removed when this value is below $\epsilon_0$, two classifiers are selected via roulette wheel with the deletion vote as described above and then the rule with the largest total number of active weights is selected.

Given the significant computational cost in training and evaluating large numbers of deep neural networks, a simple adaptive population sizing algorithm is also explored here with the aim of minimising the number of classifiers $N$ while retaining sufficient diversity throughout the search process. A moving average of the mean match set size $|[M]|$ is therefore tracked, as above for the system error. To maintain the mean $|[M]|$ within the range $[100,200]$, the maximum population size $N$ is potentially adjusted after the insertion of offspring in $[P]$, but before the population size limit is enforced. In the adaptive approach tested here, $N$ is incremented by 1 if the mean $|[M]|<100$ and $N<5000$. $N$ is decremented by 1 if the mean $|[M]|>200$ and $N>200$.

\subsection{Experimental Setup}

In all cases, hidden rectified linear units (RELUs)~\cite{He:2015} are used for activation functions. Each $cl.C$ is initialised with a single fully-connected hidden layer containing a single unit and a fully-connected output layer with a single (linear) unit. The number of units in the hidden layer may adapt, however the output layer remains fixed. Two $cl.P$ architectures are explored:

\begin{enumerate}
    \item four fully-connected hidden layers (fc)
    \item convolutional layer $\rightarrow$ maxpool layer $\rightarrow$ convolutional layer $\rightarrow$ maxpool layer $\rightarrow$ convolutional layer $\rightarrow$ fully connected layer (conv).
\end{enumerate}

All convolutional layers use a fixed $3\times3$ kernel with a stride and padding of 1. All maxpooling layers use a fixed size and stride of 2. A softmax output layer is used for $cl.P$ as previously mentioned.

The following publicly available data sets are used for initial evaluation from \url{https://www.openml.org}:
\begin{enumerate}
    \item \textsc{usps digits}: 256 features; 10 classes; 9298 instances. OpenML ID: 41082.
    \item \textsc{mnist digits}: 784 features; 10 classes; 70000 instances. OpenML ID: 554.
\end{enumerate}

For \textsc{usps digits}, the data set is divided into training and test sets using a 90--10\% split. The training set is further subdivided such that 10\% is reserved for (non-learning) validation. For \textsc{mnist digits}, the traditional 60000--10000 split for training and testing sets is used. This training set is also further subdivided to reserve 10\% for validation.

Each experiment is run for 1 million learning trials where the system alternates between performing 1000 learning trials using the training set, followed by scoring the validation set. For \textsc{usps digits}, a single pass through the entire validation set (837 instances) is used to calculate the classification error, whereas a random sample of 1000 validation instances (from the 6000 in total) are used for \textsc{mnist digits} due to the computational cost. Evaluation on test sets is performed using the population set at the minimum validation set error using a simple moving average over the most recent 10 iterations, i.e., 10000 trials. Learning parameters used may be found in Table~\ref{table:params}. Most parameters are standard defaults, e.g., \cite{Preen:2019}, however here a smaller classifier update rate $\beta$ and larger delay between EA invocations $\theta_{\text{EA}}$ are used to increase the accuracy represented over a larger window of training samples. The source code used for experimentation is available from \cite{xcsf}.

\begin{table}[t]
	\caption{Learning Parameters}
    \centering
    \begin{tabular}{l c c}
        \toprule
        Description & Parameter & Value \\
        \midrule
        Maximum population size (in micro-classifiers) & $N$ & 500 \\
        Population initialised with random classifiers & $P_{\text{init}}$ & true \\
        Target error, under which accuracy is set to 1 & $\epsilon_0$ & 0.01 \\
        Update rate for fitness, error, and set size & $\beta$ & 0.05 \\
        Accuracy offset (1=disabled) & $\alpha$ & 1 \\
        Accuracy slope & $\nu$ & 5 \\
        Fraction of classifiers to increase deletion vote & $\delta$ & 0.1 \\
        Classifier deletion threshold & $\theta_{\text{del}}$ & 100 \\
        Classifier initial fitness & $F_I$ & 0.01 \\
        Classifier initial error & $\epsilon_I$ & 0 \\
        Offspring fitness reduction (1=disabled) & $F_R$ & 0.1 \\
        Offspring error reduction (1=disabled) & $\epsilon_R$ & 1 \\ 
        Minimum number of actions in $[M]$ & $\theta_{\text{mna}}$ & 1 \\
        EA invocation frequency & $\theta_{\text{EA}}$ & 100 \\
        Number of offspring per EA invocation & $\lambda$ & 2 \\
        Crossover probability & $\chi$ & 0 \\
        Stochastic gradient descent momentum & $\omega$ & 0.9 \\
        Max.\ units added or removed per mutation & $h_M$ & \{1,5\} \\
        Whether EA subsumption is performed & \em{EASubsume} & false \\
        Whether set subsumption is performed & \em{SetSubsume} & false \\
        \bottomrule
    \end{tabular}
    \label{table:params}
\end{table}

For each experiment, we report the mean wall-clock time taken to reach 1 million learning trials (including iterative validation scoring) with an Intel\textsuperscript{\textregistered} Xeon\textsuperscript{\textregistered} CPU E5-2650 v4 @ 2.20GHz and 64GB RAM using 36 of the 48 cores.

In addition, we report the fraction of inputs matched by the single best rule $cl^{*}_{\text{mfrac}}$ as a measure of generalisation, following \cite{Preen:2019}. This rule is determined as follows. If no classifier has an error below $\epsilon_0$, the classifier with the lowest error is chosen. If more than one classifier has an error below $\epsilon_0$, the classifier that matches the largest number of inputs is used. 

The hypothesis that LCS adaptive niching can improve performance is tested by comparison with the same system, where however $cl.C$ always match $\vec{x}$. Classifier updates and the EA are thus performed within $[P]$, and single networks that cover the entire state-space are designed. This configuration operates as a more traditional EA, and we henceforth refer to this system as the EA. The EA acts as a control with which we can make direct comparison since it operates with the same evolutionary and gradient descent operators, and uses the same parameters. Any future changes or improvements made to the EA may be incorporated within XCSF.


In addition, we compare performance with single networks of identical architecture using Keras/Tensorflow~\cite{keras}. Test scores are reported from using the weights from the lowest validation error over an equivalent number of trials. For the closest comparison, stochastic gradient descent using a learning rate of 0.001 and batch size of 1 are used.

\section{Results}
\label{section:results}

\begin{figure*}[t]
    \subfloat[Four fully-connected prediction hidden layers without connection mutation; maximum growth rate $h_M=1$.]{%
        \includegraphics[width=\textwidth]{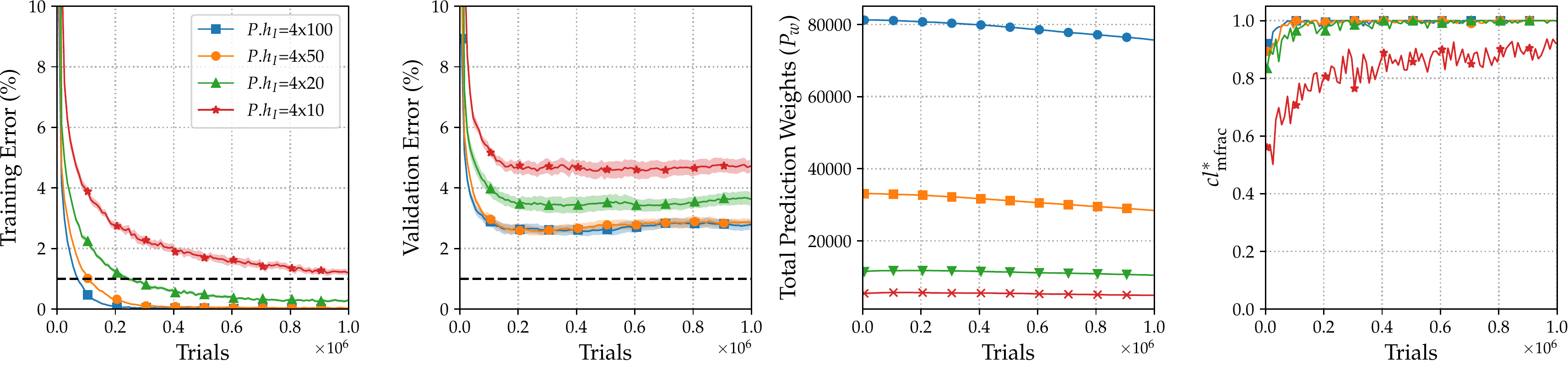}}
    \hfill
    \subfloat[Four fully-connected prediction hidden layers with connection mutation; maximum growth rate $h_M=1$.]{%
        \includegraphics[width=\textwidth]{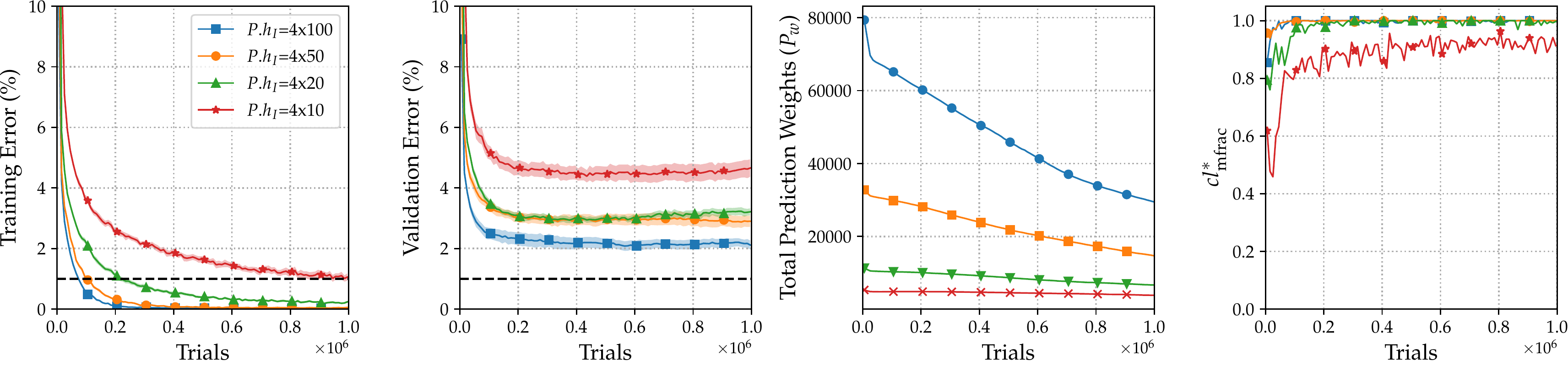}}
    \hfill
    \subfloat[Four fully-connected prediction hidden layers with adaptive population sizing and connection mutation; maximum growth rate $h_M=1$.]{%
        \includegraphics[width=\textwidth]{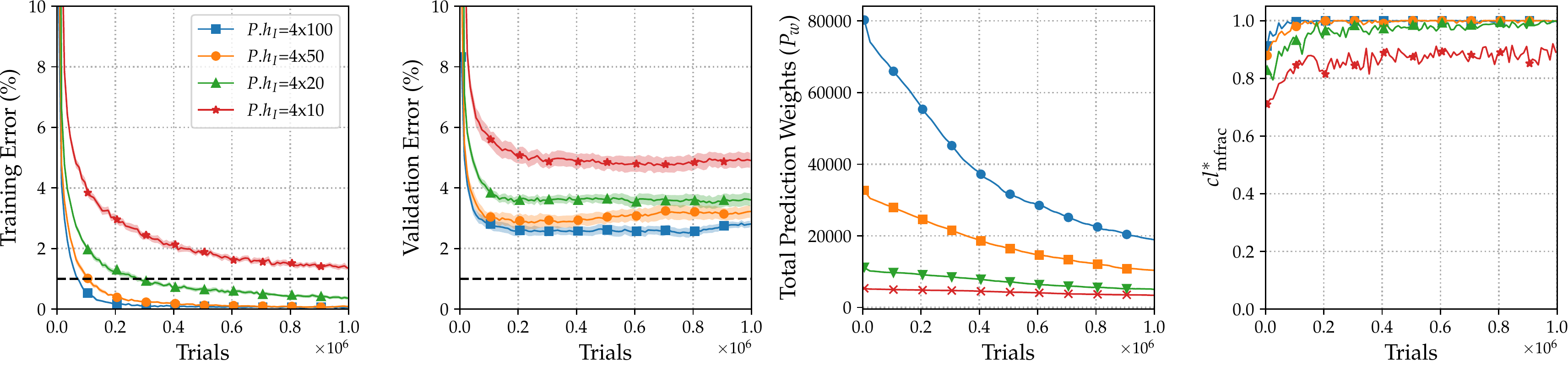}}    
    \hfill
    \subfloat[Convolutional prediction layers with connection mutation; initial number of units $h_I=1$; maximum growth rate $h_M \in \{1,5\}$.]{%
        \includegraphics[width=\textwidth]{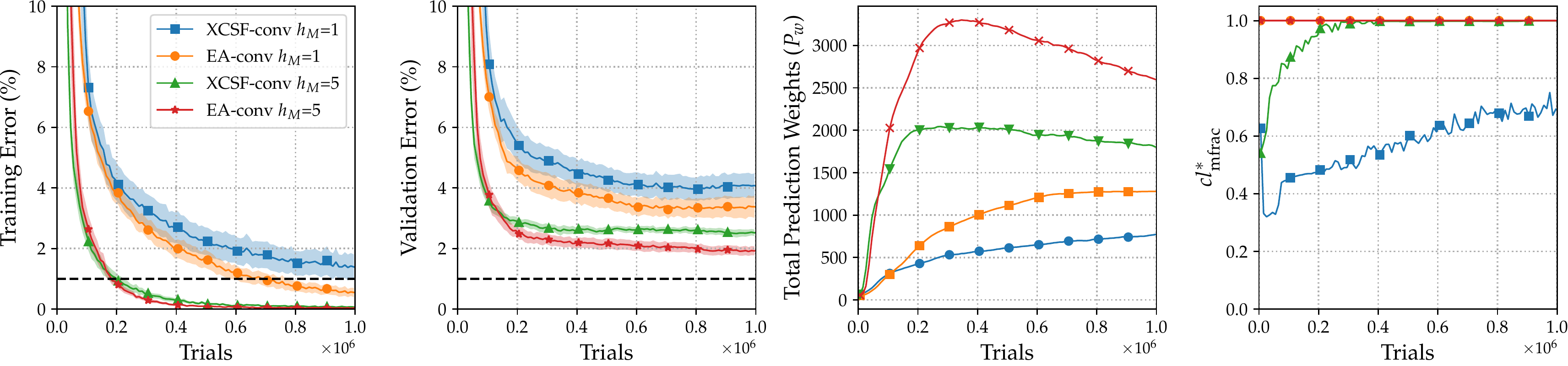}}            
        
    \caption{\textsc{usps digits}. Shown are the training and validation errors (standard error shaded), the total number of prediction weights ($P_w$) and the fraction of inputs matched by the best rule ($cl^{*}_{\text{mfrac}}$). Mean population set values reported.}
    \label{fig:usps}
\end{figure*}

\begin{figure*}[t]
    \subfloat[Four fully-connected prediction hidden layers without connection mutation; maximum growth rate $h_M=1$.]{%
        \includegraphics[width=\textwidth]{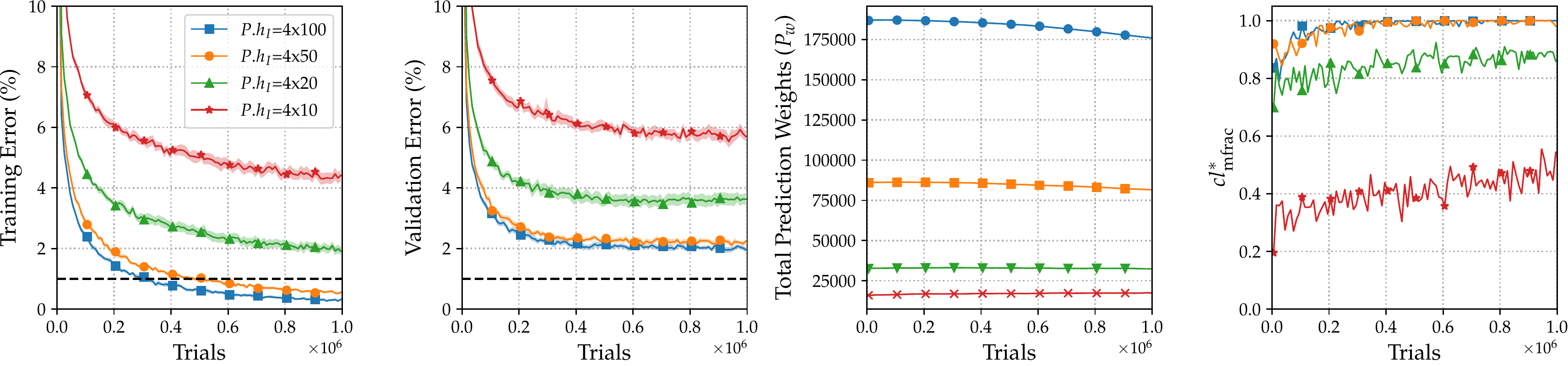}}
    \hfill
    \subfloat[Four fully-connected prediction hidden layers with connection mutation; maximum growth rate $h_M=1$.]{%
        \includegraphics[width=\textwidth]{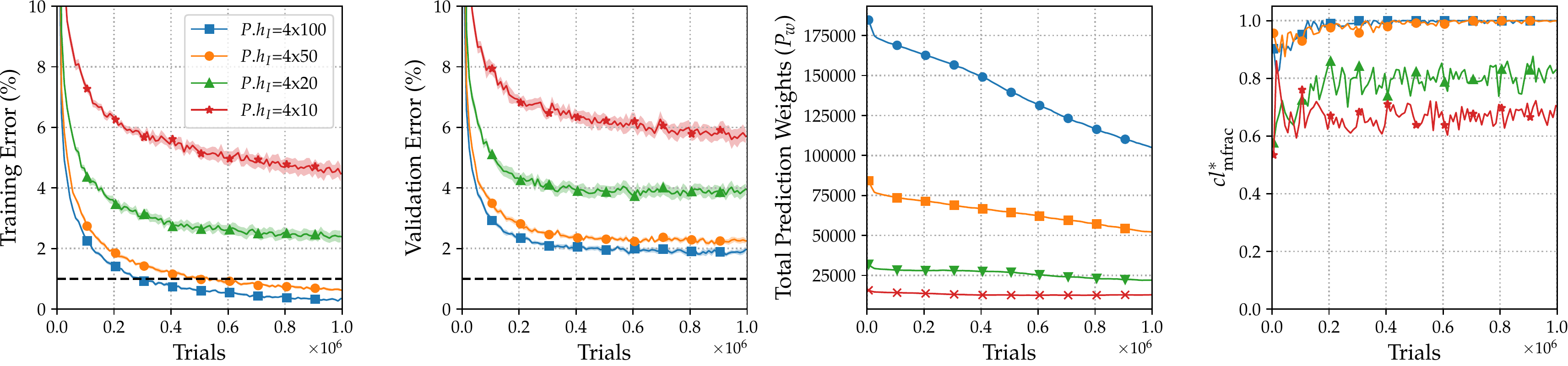}}
    \hfill
    \subfloat[Four fully-connected prediction hidden layers with adaptive population sizing and connection mutation; maximum growth rate $h_M=1$.]{%
        \includegraphics[width=\textwidth]{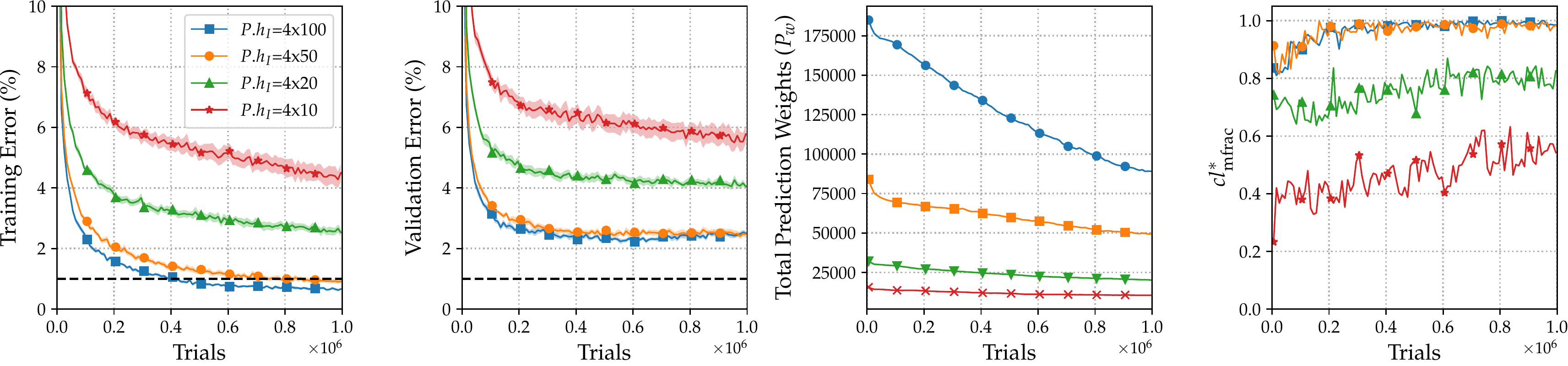}}    
    \hfill
    \subfloat[Convolutional prediction layers with connection mutation; initial number of units $h_I=1$; maximum growth rate $h_M \in \{1,5\}$.]{%
        \includegraphics[width=\textwidth]{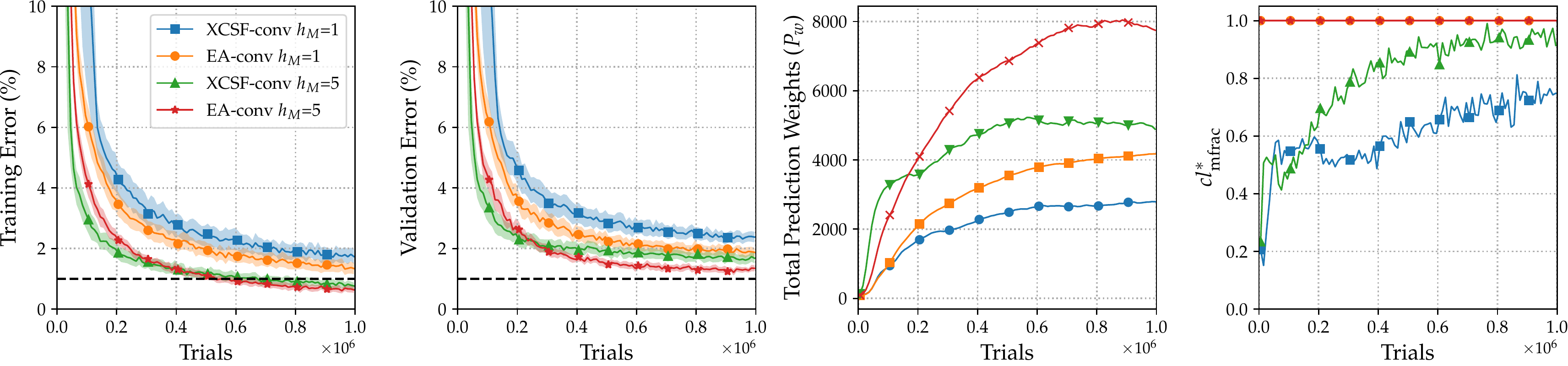}}          
    \caption{\textsc{mnist digits}: Shown are the training and validation errors (standard error shaded), the total number of prediction weights ($P_w$) and the fraction of inputs matched by the best rule ($cl^{*}_{\text{mfrac}}$). Mean population set values reported.}
    \label{fig:mnist}
\end{figure*}

Fig.~\ref{fig:usps} and Fig.~\ref{fig:mnist} illustrate the performance on the \textsc{usps digits} and \textsc{mnist digits} data sets. Initial number of hidden units $h_I$. Table~\ref{table:summary} presents a summary of the test results. Similar to results widely reported elsewhere, convolutional layers are observed to produce better accuracy scores than connected layers within XCSF and the EA.

The adaptive population sizing algorithm significantly reduced wall-clock time. For example, by an average 53\% on \textsc{mnist digits} with 100 initial neurons in each connected layer. However, it appears to have resulted in a small reduction in mean classification accuracy, e.g., 97.69\% adaptive vs.\ 98.11\% non-adaptive in the same experiment. This suggests that the $[100,200]$ range used to maintain the mean $|[M]|$ may be too small.

While XCSF identifies suitable solutions with very low training error, its generalisation scores on \textsc{mnist digits} test accuracy ($98.37\pm0.15$\%) are not significantly better than the EA ($98.53\pm0.11$\%), which evaluates all candidates as global models. As further reference, linear regression~\cite{LeCun:1998} achieves $91.6$\% accuracy. Haar-like feature conditions within XCS~\cite{Ebadi:2014} scored $91\pm1$\% (increased to $95$\% with rotation), and $94\pm1$\% within UCS~\cite{Ebadi:2012}; both were executed for 4 million generations (54000--74000$s$ wall-clock time).

\begin{table*}[t]
    \caption{Summary of test score results. Shown are the algorithm, the model type, the maximum growth rate ($h_M$), the population size $N$, the initial number of prediction hidden units in each layer ($P.h_I$), whether connection mutation was applied, the mean test accuracy $\pm$ the standard error (\%), the median and maximum test accuracy, the mean wall-clock time to 1 million trials, and the mean number of trials at which the lowest validation error moving average was recorded ($[P]^*_{\textnormal{trials}}$). Row with smallest mean shaded.}
    \centering
    \begin{tabular}{l l r r r l c c c r r}
        \toprule
        Algorithm & Model Type & $h_M$ & $N$ & $P.h_I$ & Connect. Mut. & Mean $\pm$ SE & Median & Max & Time (s) & $[P]^*_{\text{trials}}$\\
        \midrule
        \underline{\textsc{usps digits}} &&&&&&&&&& \\
        XCSF & Connected & 1 & 500 & 100 & Disabled & 97.29 $\pm$ 0.14 & 97.31 & 97.96 & 29694 & 438200\\
        XCSF & Connected & 1 & 500 & 50 & Disabled & 97.22 $\pm$ 0.18 & 97.10 & 98.17 & 11061 & 492500\\
        XCSF & Connected & 1 & 500 & 20 & Disabled & 97.13 $\pm$ 0.17 & 97.20 & 97.85 & 1564 & 427000\\
        XCSF & Connected & 1 & 500 & 10 & Disabled & 94.85 $\pm$ 0.43 & 95.16 & 96.67 & 502 & 532300\\

        XCSF & Connected & 1 & 500 & 100 & Enabled & 97.20 $\pm$ 0.17 & 97.48 & 98.39 & 30828 & 542100\\
        XCSF & Connected & 1 & 500 & 50 & Enabled & 97.57 $\pm$ 0.08 & 97.58 & 98.17 & 11744 & 473700\\
        XCSF & Connected & 1 & 500 & 20 & Enabled & 96.62 $\pm$ 0.19 & 96.45 & 97.74 & 1558 & 606500\\
        XCSF & Connected & 1 & 500 & 10 & Enabled & 95.68 $\pm$ 0.16 & 95.70 & 96.56 & 590 & 572400\\

        XCSF & Connected & 1 & Adaptive & 100 & Enabled & 97.41 $\pm$ 0.14 & 97.36 & 98.17 & 13080 & 659100\\
        XCSF & Connected & 1 & Adaptive & 50 & Enabled & 97.07 $\pm$ 0.13 & 97.04 & 97.63 & 2681 & 498800\\
        XCSF & Connected & 1 & Adaptive & 20 & Enabled & 96.36 $\pm$ 0.22 & 96.29 & 97.53 & 552 & 476300\\
        XCSF & Connected & 1 & Adaptive & 10 & Enabled & 95.48 $\pm$ 0.25 & 95.81 & 96.56 & 434 & 515000\\

        Keras SGD & Connected & n/a & 1 & 100 & n/a & 97.42 $\pm$ 0.16 & 97.42 & 98.28 & n/a & n/a\\
        Keras SGD & Connected & n/a & 1 & 50 & n/a & 97.17 $\pm$ 0.14 & 97.20 & 97.74 & n/a & n/a\\
        Keras SGD & Connected & n/a & 1 & 20 & n/a & 96.10 $\pm$ 0.17 & 95.91 & 97.10 & n/a & n/a\\
        Keras SGD & Connected & n/a & 1 & 10 & n/a & 94.22 $\pm$ 0.24 & 94.36 & 95.16 & n/a & n/a\\

        XCSF & Convolutional & 1 & 500 & 1 & Enabled & 95.84 $\pm$ 0.50 & 96.24 & 97.31 & 5745 & 761500\\
        EA & Convolutional & 1 & 500 & 1 & Enabled & 96.74 $\pm$ 0.21 & 96.72 & 97.85 & 6768 & 815400\\
        XCSF & Convolutional & 5 & 500 & 1 & Enabled & 97.68 $\pm$ 0.19 & 97.58 & 98.60 & 3534 & 755500\\
        \rowcolor[HTML]{EFEFEF}
        EA & Convolutional & 5 & 500 & 1 & Enabled & 98.10 $\pm$ 0.20 & 97.91 & 99.25 & 6919 & 811800\\
        
        \midrule
        \underline{\textsc{mnist digits}} &&&&&&&&&& \\
        XCSF & Connected & 1 & 500 & 100 & Disabled & 97.92 $\pm$ 0.09 & 98.00 & 98.24 & 57651 & 711400 \\
        XCSF & Connected & 1 & 500 & 50 & Disabled & 97.70 $\pm$ 0.07 & 97.70 & 98.10 & 24607 & 723100 \\
        XCSF & Connected & 1 & 500 & 20 & Disabled & 96.26 $\pm$ 0.24 & 96.42 & 97.10 & 7454 & 649300 \\
        XCSF & Connected & 1 & 500 & 10 & Disabled & 94.35 $\pm$ 0.13 & 94.32 & 95.11 & 867 & 756200 \\
        
        XCSF & Connected & 1 & 500 & 100 & Enabled & 98.11 $\pm$ 0.08 & 98.19 & 98.36 & 62955 & 858200\\
        XCSF & Connected & 1 & 500 & 50 & Enabled & 97.68 $\pm$ 0.05 & 97.67 & 97.90 & 26095 & 725500\\
        XCSF & Connected & 1 & 500 & 20 & Enabled & 96.07 $\pm$ 0.15 & 96.23 & 96.57 & 5021 & 606300\\
        XCSF & Connected & 1 & 500 & 10 & Enabled & 94.28 $\pm$ 0.25 & 94.32 & 95.51 & 3910 & 869700\\

        XCSF & Connected & 1 & Adaptive & 100 & Enabled & 97.69 $\pm$ 0.07 & 97.75 & 97.93 & 27025 & 599200\\
        XCSF & Connected & 1 & Adaptive & 50 & Enabled & 97.37 $\pm$ 0.08 & 97.40 & 97.64 & 11926 & 649300\\
        XCSF & Connected & 1 & Adaptive & 20 & Enabled & 95.76 $\pm$ 0.18 & 95.81 & 96.49 & 1798 & 752700\\
        XCSF & Connected & 1 & Adaptive & 10 & Enabled & 94.13 $\pm$ 0.28 & 94.07 & 95.47 & 1015 & 883400\\

        Keras SGD & Connected & n/a & 1 & 100 & n/a & 97.79 $\pm$ 0.05 & 97.79 & 98.06 & n/a & n/a\\
        Keras SGD & Connected & n/a & 1 & 50 & n/a & 97.22 $\pm$ 0.04 & 97.19 & 97.75 & n/a & n/a\\
        Keras SGD & Connected & n/a & 1 & 20 & n/a & 95.48 $\pm$ 0.10 & 95.50 & 96.00 & n/a & n/a\\
        Keras SGD & Connected & n/a & 1 & 10 & n/a & 92.74 $\pm$ 0.24 & 92.83 & 93.76 & n/a & n/a\\

        XCSF & Convolutional & 1 & 500 & 1 & Enabled & 97.58 $\pm$ 0.19 & 97.85 & 98.17 & 20349 & 909600\\     
        EA & Convolutional & 1 & 500 & 1 & Enabled & 97.98 $\pm$ 0.18 & 98.05 & 98.72 & 27873 & 873100\\    
        XCSF & Convolutional & 5 & 500 & 1 & Enabled & 98.37 $\pm$ 0.15 & 98.49 & 98.91 & 13426 & 770700\\
        \rowcolor[HTML]{EFEFEF}
        EA & Convolutional & 5 & 500 & 1 & Enabled & 98.53 $\pm$ 0.11 & 98.61 & 98.89 & 41080 & 844600\\    
        \bottomrule
    \end{tabular}
    \label{table:summary}
\end{table*}

\begin{figure*}[t]
    \subfloat[\textsc{usps digits}]{%
        \includegraphics[width=\textwidth]{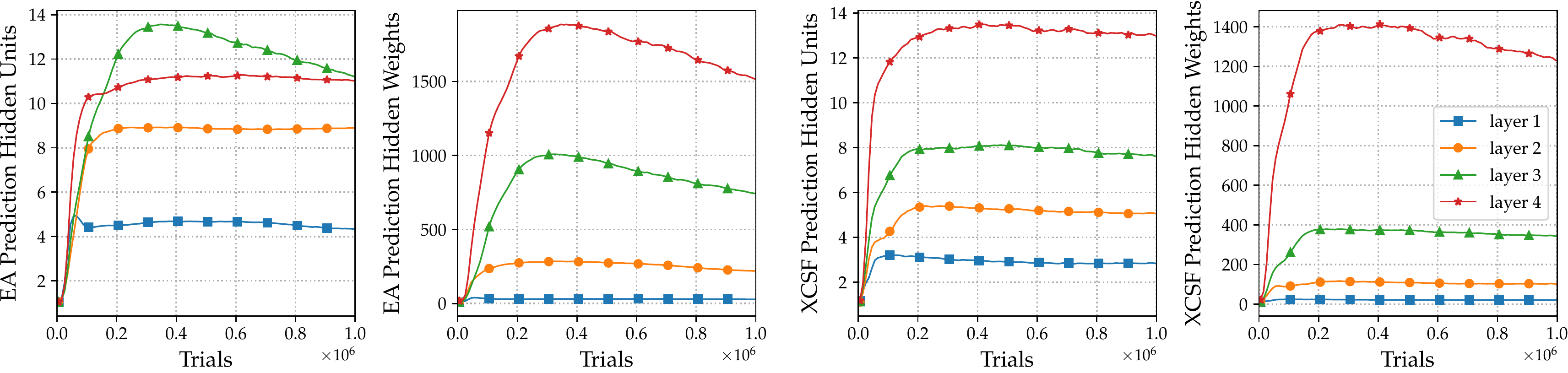}}
    \hfill        
    \subfloat[\textsc{mnist digits}]{%
        \includegraphics[width=\textwidth]{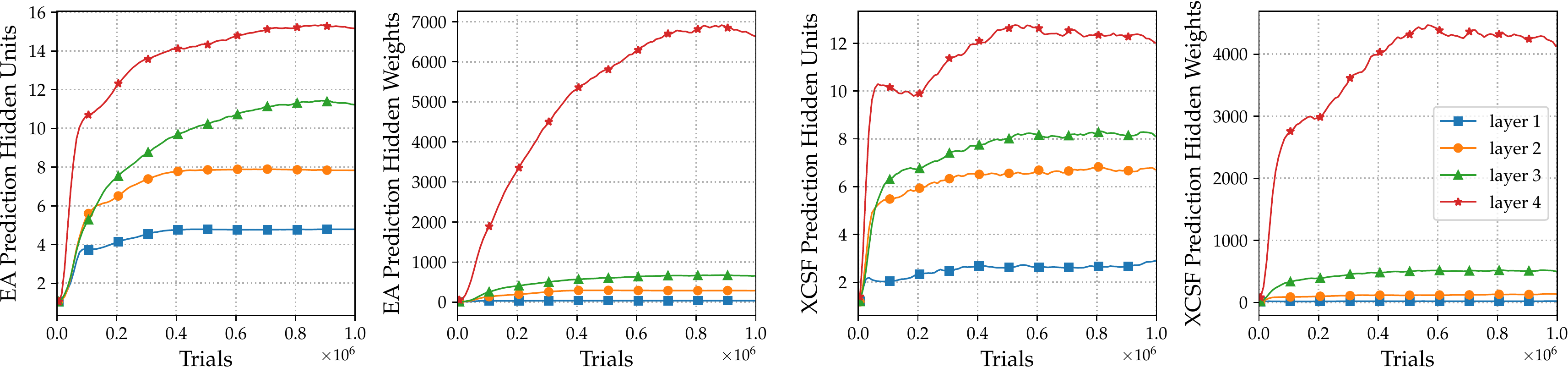}}
    \caption{Topology of convolutional prediction layers with $h_M=5$. Shown are the number of EA hidden units and weights in each layer; and the number of XCSF hidden units and weights in each layer. Mean population set values reported.}
    \label{fig:mnist_conv}
\end{figure*}

\section{Conclusion}
\label{section:conclusion}

This article has presented the first results from using a deep neural network representation within XCSF. Both fully-connected and convolutional networks were explored on handwritten digit recognition tasks. Current work is exploring ways in which to improve upon these results and provide benchmark data for a wider range of data sets, and applications. 



\end{document}